\documentclass{article}

 \usepackage[preprint, nonatbib]{neurips_2020}

\usepackage[utf8]{inputenc} % allow utf-8 input
\usepackage[T1]{fontenc}    % use 8-bit T1 fonts
\usepackage{hyperref}       % hyperlinks
\usepackage{url}            % simple URL typesetting
\usepackage{booktabs}       % professional-quality tables
\usepackage{amsfonts}       % blackboard math symbols
\usepackage{nicefrac}       % compact symbols for 1/2, etc.
\usepackage{microtype}      % microtypography
\usepackage{graphicx}
\usepackage{amsmath}
\usepackage{biblatex}
\title{Higher Order Linear Transformer}

\author{%
  Jean Mercat\\
  Paris Saclay University\\
  Laboratory of Signal and System (L2S)\\
  Gif sur Yvette, \\
  \texttt{jean.mercat@centralesupelec.fr} \\
}

\begin{document}

\maketitle

\begin{abstract}
  Following up the linear transformer part of the article~\cite{Katharopoulos2020}, that takes this idea from~\cite{Shen2018}, the trick that produces a linear complexity for the attention mechanism is re-used and extended to a second order approximation of the softmax normalization.
\end{abstract}

\section{Introduction}
I cannot write a better introduction than the one from~\cite{Katharopoulos2020} and since copy-pasting is frowned upon, you will have to suffer a click on this link: \url{https://arxiv.org/pdf/2006.16236.pdf}.

\section{Related Work}
Same thing same link: \url{https://arxiv.org/pdf/2006.16236.pdf}.\\
However, we can add that the results from~\cite{Katharopoulos2020} do not quiet catch up with the softmax normalization attention. 
This motivates an improvement that would match the original attention results with a linear complexity.
However, the result improvement is not a claim because we only tested our model on random data.

\section{Taylor Expansion of Exponential}

\begin{figure}
    \centering
    \includegraphics[width=\textwidth]{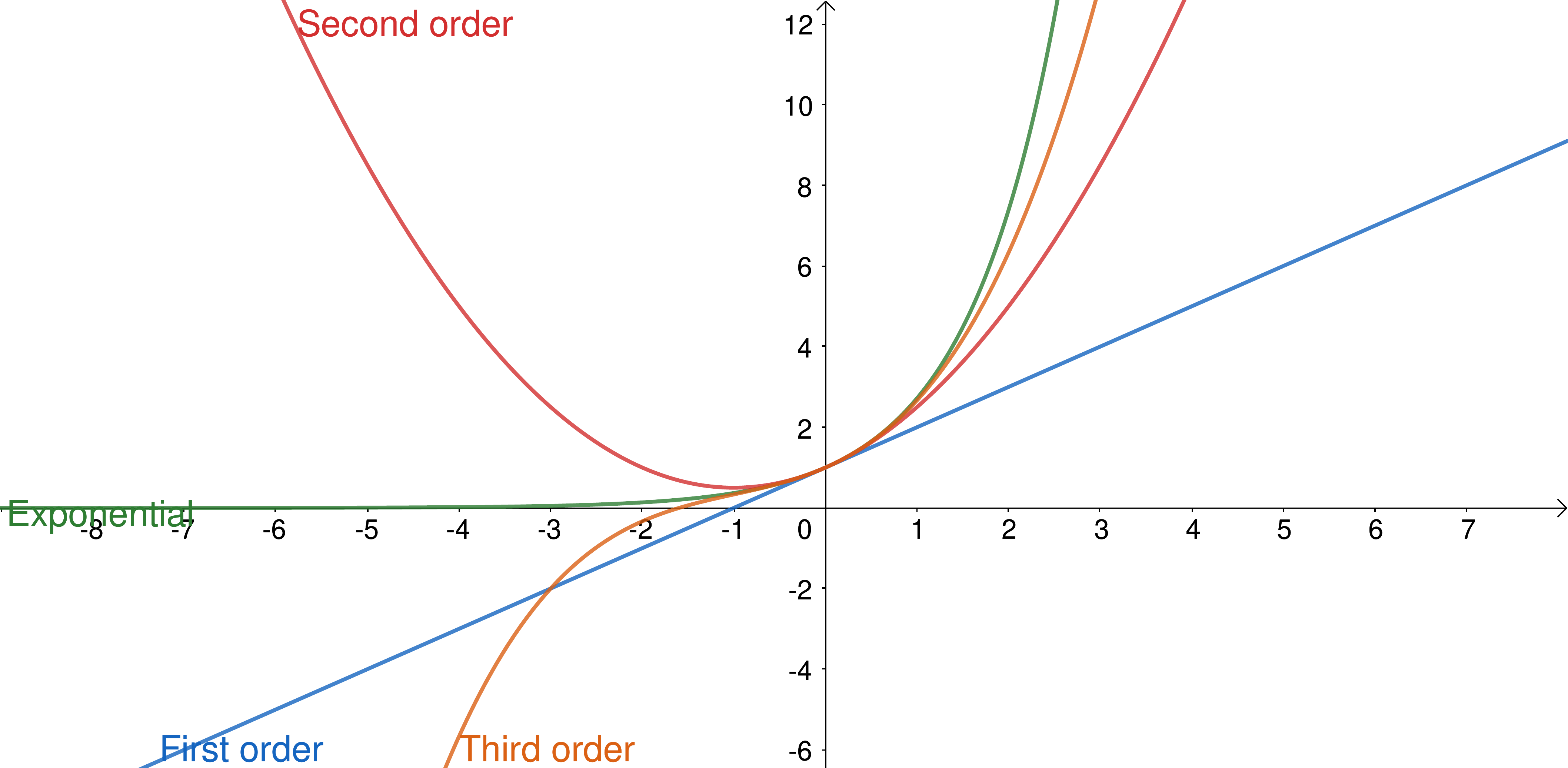}
    \caption{First, second and third order Taylor expansion of exponential}
    \label{fig:taylor}
\end{figure}

We want to approximate $\operatorname{Softmax}(Q K^T)$.
$\operatorname{Softmax}(x) = \frac{\exp(x)}{\sum \exp(x)}$
Thus, approximating the $\operatorname{Softmax}$ involve approximating the exponential function.
We use the second order Taylor expansion $\exp(x) \approx 1 + x + \frac{x^2}{2} + ...$ for small $x$.
Well, nothing new here but a nearby visual is always welcome.
See figure~\ref{fig:taylor} that represent the exponential, $x \rightarrow 1 + x$, $x \rightarrow 1 + x + \frac{x^2}{2}$ and $x \rightarrow 1 + x + \frac{x^2}{2} + \frac{x^3}{6}$.
We see that the approximation is quickly very wrong when the values are not close to 0.
Thus, the values of the matrix $Q K^T$ must remain around 0.
To that end, the values of $Q$ and $K$ are normalized with a layer normalization~\cite{Ba2016} without the element-wise affine rescaling.
Moreover, it is rescaled with the vector dimension $d$ as done in~\cite{Vaswani2017} but to keep closer to 0, we add a parameter $\alpha > 1$.
I chose $\alpha=3$ and an expansion to the second order.
Thus, we approximate the exponential of the terms of the following matrix: \[\frac{\tilde{Q} \tilde{K}^T}{\alpha\sqrt{d}}\]
With, $\tilde{Q} = \operatorname{LayerNorm}(Q)$ and $\tilde{K} = \operatorname{LayerNorm}(K)$.

The second order Taylor expansion approximation is given by:
\begin{equation}
    \exp\left(\frac{\tilde{Q} \tilde{K}^T}{\alpha\sqrt{d}}\right) \approx 1 + \frac{\tilde{Q} \tilde{K}^T}{\alpha\sqrt{d}} + \frac{(\frac{\tilde{Q} \tilde{K}^T}{\alpha\sqrt{d}})^2}{2}
    \label{eq:taylor}
\end{equation}
Where the square exponent is to be performed element-wise.

This paragraph is made of intuitive far fetched statements.
If this kind on non-scientific remarks gets to your nerves, please skip ahead to the next section.
I expect that using higher order expansions would allow lower $\alpha$ values that would in turn help centering the distribution of gradients through this operation.
The even orders overestimate greatly the function for negative antecedent and the odd orders greatly underestimate it.
This means that with even orders, keys and queries that are aligned in opposite directions produce the about same effect as the one aligned in the same direction.
With a large $d$ dimension, the model can probably find orthogonal vectors so it produces a 0 values but this means that the correlated vectors must compensate the 1 value produced by the exponential of 0.
Thus, we consider subtracting one to this expression in order for the 0 correlation to be achievable without needing to compensate with very high correlation values. 
With odd orders, the opposite effect is produced.
Negative values could cancel out the positive values in the norm computation which is a weird behavior.
Thus, we want to use even orders and 2 is the simplest non-trivial one.

\section{Efficient second order normalization}

The attention computation that we are interested in is written as follows: $\operatorname{Softmax}(Q K^T)V$

The trick used for efficiency in~\cite{Shen2018} is to re-order the computation in order to sum the elements along the large dimension first: $\left(QK^T\right)V = Q\left(K^T V\right)$.
If there are $n$ elements, the attention matrix is of size $n \times n$.
To reduce memory usage, it should not be computed explicitly.
However, this cannot be done with the softmax operator.
Thus, a normalization function $\rho$ that can be distributed is used instead of the softmax: $\left(\rho(Q)\rho(K)^T\right)V = \rho(Q)\left(\rho(K)^T V\right)$.

We use the same trick with the expression~\eqref{eq:taylor} such that the normalization function approximates the softmax normalization.
\begin{equation}
    \exp\left(\frac{\tilde{Q} \tilde{K}^T}{\alpha\sqrt{d}}\right)V \approx \left(1 + \frac{\tilde{Q} \tilde{K}^T}{\alpha\sqrt{d}} + \frac{(\frac{\tilde{Q} \tilde{K}^T}{\alpha\sqrt{d}})^2}{2}\right)V
    \label{eq:attention1}
\end{equation}
The first two terms can be computed efficiently directly.
The third term must be re-written using the multinomial expansion: 
\begin{equation}
    \begin{split}
        \left((\tilde{Q} \tilde{K}^T)^2 V\right)_i &= \sum_{j=1}^n \left(\sum_{m=1}^{d}q_i^m k_j^m \right)^2 \mathbf{v}_j\\
        &= \sum_{m, l} q_i^m q_i^l \sum_{j=1}^n k_j^m k_j^l \mathbf{v}_j
    \end{split}
\end{equation}
Of course, this can be generalized with higher order Taylor expansions with the same idea.
However, with this formulation, the complexity is $n d_v d_k^{o_t}$ with $o_t$ the order of the Taylor expansion $d_v$ the dimension of the value vectors and $d_k$ the dimension of the key and query vectors.
Thus, it is unlikely that the benefit of higher order expansion would both ensure $n d_v d_k^{o_t} < n^2 d_v d_k$ and improve the results.
The normalization term $\sum_{j=1}^n \exp\left(\frac{\tilde{Q} \tilde{K}^T}{\alpha\sqrt{d}}\right)$ is approximated and computed in the same fashion.
The code forked from~\cite{Katharopoulos2020} is accessible on GitHub \url{https://github.com/jmercat/fast-transformers/tree/linear-softmax} but is yet to be tested on real data.

\section{Application}

I will do it if time allows it.
If you do it before me, feel free to write this section and your name on the first page.

\section*{Broader Impact}

This work will revolutionize the universe in 3 simple steps: Firstly, GPT-4 implements our proposition to use a larger context window. Then, its 100 trillion parameters are trained for one year on all the TPUs. Finally, the resulting model can be asked to improve itself. This initiates the singularity that will transform the world as we know it and therefore should be used with care. Especially because GPT-4 will be as sexist, as racist and as all -ist words combined as the context it was trained on (thus, very much).

\begin{ack}
This «feuille de chou» is brought to you by the procrastination of its author that should be writing his memoir instead. The NeurIPS template is used to make it look serious but as you have come to understand, this is not so serious.
\end{ack}

\medskip
\small

%\printbibliography

\end{document}